\documentclass{article}

 \usepackage[preprint]{neurips_2026}

\usepackage[utf8]{inputenc} 
\usepackage[T1]{fontenc}    
\usepackage{hyperref}       
\usepackage{url}            
\usepackage{booktabs}       
\usepackage{amsfonts}       
\usepackage{amsmath}        
\usepackage{amssymb}        
\usepackage{nicefrac}       
\usepackage{microtype}      
\usepackage{xcolor}         
\usepackage{algorithm}
\usepackage{algorithmic}
\usepackage{graphicx}

\newcommand{\x}{\mathbf{x}}
\newcommand{\vtheta}{v_\theta}

\newcommand{\Gcal}{\mathcal{G}}
\newcommand{\Scal}{\mathcal{S}}

\usepackage{tabularx}
\newcolumntype{C}{>{\centering\arraybackslash}X}
\usepackage{lipsum}
\usepackage{subcaption}
\newcommand{\supurl}{https://ernestchu.github.io/hsa}

\title{Not All Tokens Need 40 Steps: Heterogeneous Step Allocation in Diffusion Transformers for Efficient Video Generation}

\author{%
  Ernie Chu \\
  Johns Hopkins University\\
  Baltimore, MD 21218 \\
  \texttt{schu23@jhu.edu} \\
  \And
  Vishal M. Patel \\
  Johns Hopkins University\\
  Baltimore, MD 21218 \\
  \texttt{vpatel36@jhu.edu} \\
}

\begin{document}

\maketitle

\begin{abstract}
Diffusion Transformers (DiTs) have achieved state-of-the-art video generation quality, but they incur immense computational cost because standard inference applies the same number of denoising steps uniformly to every token in the sequence. It is well known that human vision ignores vast amounts of redundant motion. Why, then, do our densest models treat every spatiotemporal token with equal priority? In this paper, we introduce Heterogeneous Step Allocation (HSA), a training-free inference algorithm that assigns varying step budgets to different spatiotemporal tokens based on their velocity dynamics. To resolve the resulting sequence-length mismatch without sacrificing global context, HSA introduces a KV-cache synchronization mechanism that allows active tokens to attend to the full sequence while entirely bypassing inactive tokens. Furthermore, we derive a cached Euler update that advances the latent states of skipped tokens in a single operation without additional model evaluations. We evaluate HSA on the Wan-2 and LTX-2 models for both text-to-video (T2V) and image-to-video (I2V) generation. Our results demonstrate that HSA significantly outperforms previous state-of-the-art caching methods and the vanilla Flow Matching baseline, especially at aggressive acceleration regimes (e.g., 50\% and 25\% runtimes). Crucially, HSA achieves a superior quality-runtime Pareto frontier without the need for expensive offline profiling, robustly preserving structural integrity and generation quality even under tight computational budgets.

Project page: \url{\supurl}

\end{abstract}

\section{Introduction}

Diffusion Transformers (DiTs)~\cite{peebles2023dit} have rapidly emerged as the architecture of choice for high-fidelity generative modeling across image, video, and audio domains~\cite{esser2024scaling,labs2025flux,ma2024latte,opensora2024,kong2024hunyuanvideo,yang2025cogvideox,wan2025,ltx2026}. By coupling the global expressivity of transformer self-attention with iterative denoising, DiTs have achieved state-of-the-art generation quality on a variety of benchmarks. Yet this quality comes at a steep computational cost: generating a single high-resolution video clip can require dozens of full forward passes through a model with billions of parameters, each pass operating over thousands of spatiotemporal tokens.

A fundamental, yet underexplored, inefficiency lies in the \emph{uniformity} of the standard inference protocol. Every token in the sequence traverses an identical number of reverse diffusion steps. This monolithic schedule ignores a well-established property of visual data: content is highly \emph{asymmetric}. Homogeneous background regions, temporally static patches, and coarsely structured content are perceptually simpler and require much smaller number of denoising steps than detail-rich foreground objects, fine textures, or regions of high motion. Applying the full denoising budget uniformly to every token is therefore wasteful. Moreover, this rigid approach fails to account for the biases of the human visual system; viewers are significantly more sensitive to quality degradation in static video components than in highly dynamic ones~\cite{lin2013visual,lin2014fvqa}, further emphasizing the need for a perceptually optimized, non-uniform strategy across the spatiotemporal token sequence.

In this paper, we introduce \textbf{Heterogeneous Step Allocation (HSA)}, a training-free inference paradigm that assigns token-specific denoising trajectories within a pretrained DiT. Rather than forcing every token through $T$ reverse steps, HSA dynamically partitions the token sequence into groups based on their velocity dynamics. Each group is allocated a (typically smaller) number of steps that divides $T$ evenly. Tokens assigned fewer steps are updated less frequently, while a subset of \emph{baseline} tokens retains the full denoising schedule and serves as the anchor trajectory.

However, this design immediately raises a synchronization challenge: self-attention requires all tokens to attend to one another, yet tokens with heterogeneous schedules are, in general, at \emph{different noise levels} at any given wall-clock iteration. We resolve this with a lightweight \textbf{KV-cache synchronization} mechanism. At each iteration, every active token computes fresh key and value projections that are written into a per-layer cache. Active tokens then attend against the \emph{full} $N$-token cache---covering both their own freshly computed entries and the stale-but-valid entries of currently skipped tokens---preserving global receptive field at a reduced $O(|\mathcal{A}_i| \cdot N)$ attention cost. Inactive tokens are bypassed entirely, incurring no query computation, no cross-attention, and no feed-forward at that iteration.

The second challenge is updating the latent state of skipped tokens without a new model evaluation. Under the framework of Flow Matching~\cite{lipman2023flow,liu2023flow}, we address this with a \textbf{cached Euler update}: each token stores the velocity predicted at its most recent active step, and at every global iteration all tokens---active and skipped alike---are advanced by the same incremental Euler step $(\sigma_{i+1} - \sigma_i)$, with active tokens using their freshly computed velocity and skipped tokens reusing the cached one. This update is a single tensor operation over all $N$ tokens with no branching, keeping the implementation simple and GPU-friendly.

We quantitatively evaluate HSA on Wan-2.1-1.3B~\cite{wan2025}, an efficient open-source video DiT. Without any fine-tuning or time-consuming offline profiling, we show that HSA achieves a superior quality-runtime Pareto frontier compared to uniform Flow Matching and recent state-of-the-art caching methods. Its advantages are particularly pronounced at aggressive acceleration regimes (e.g., 50\% and 25\% runtimes), robustly tracking the full-budget reference across diverse evaluation dimensions where baselines suffer from catastrophic collapse. We also provide visual comparisons of larger models on our \href{\supurl}{project page}, including Wan-2.1-14B/2.2-A14B~\cite{wan2025} and LTX-2~\cite{ltx2026} audio-video generator, to show the versatility of HSA across model scales and modalities. Notably, HSA is a flexible framework that can be instantiated with a variety of token-grouping strategies, and we find that dynamic token selection consistently yields strong performance. With more sophisticated grouping strategies, we anticipate a greater potential for improvement. This paper lays the groundwork for HSA and contributes in the following ways:
\begin{itemize}
    \item We propose HSA, a fully training-free, model-agnostic, plug-and-play inference algorithm that assigns heterogeneous denoising trajectories to spatiotemporal token groups.
    \item We introduce a \textbf{KV-cache synchronization} mechanism (Sec.~\ref{sec:kv-cache-sync}) that preserves full $N$-token attention context for active tokens while bypassing inactive tokens entirely, reducing per-iteration attention cost to $O(|\mathcal{A}_i| \cdot N)$.
    \item We show that a simple \textbf{cached Euler update} (Sec.~\ref{sec:cached-euler}) that advances skipped tokens without additional model evaluations is sufficient to maintain generation quality under heterogeneous schedules.
    \item We study four different token-grouping strategies and four budget presets to give a comprehensive picture of the HSA design space, demonstrating its robustness and high visual fidelity even at tight inference budgets. (Sec.~\ref{sec:eval-setup} and Sec.~\ref{sec:results})
\end{itemize}

\begin{figure}[t]
\centering
\includegraphics[width=\linewidth]{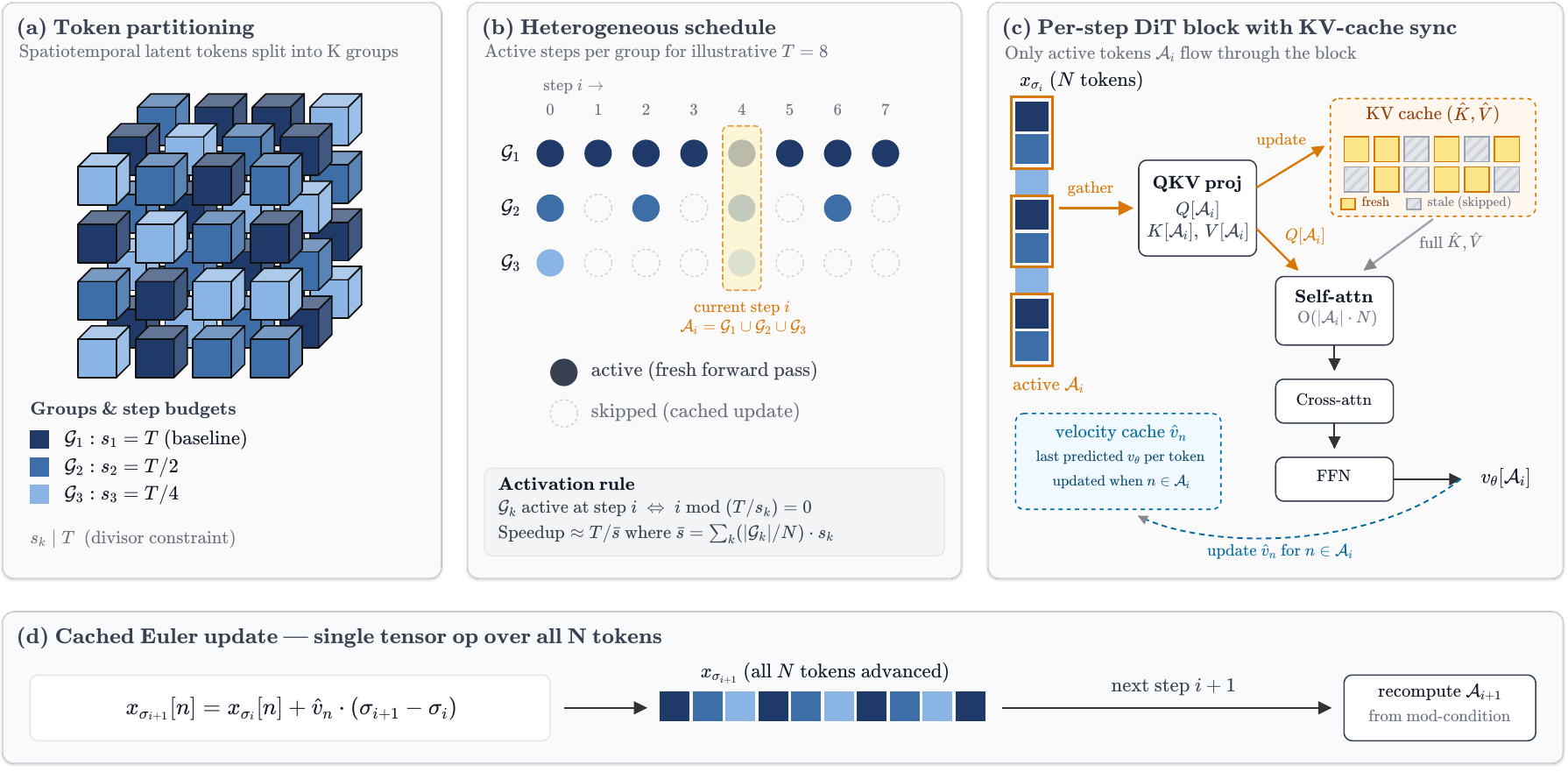}
\caption{\textbf{Overview of Heterogeneous Step Allocation (HSA).} (a) Spatiotemporal latent tokens are partitioned into $K$ disjoint groups $\Gcal_1,\ldots,\Gcal_K$, each assigned a step budget $s_k$ (a divisor of $T$), with $\Gcal_1$ as the full-budget baseline group. (b) Each group is denoised on its own schedule, yielding an active set $\mathcal{A}_i = \bigcup \{\Gcal_k : i \bmod (T/s_k){=}0\}$ (c) Per-step DiT block with KV-cache synchronization: only the active tokens $\mathcal{A}_i$ flow through QKV projection and attention, their fresh $K,V$ entries overwrite the cache, and self-attention attends active queries against the full $K,V$ (fresh + cached), reducing the per-step cost from $O(N^2)$ to $O(|\mathcal{A}_i| \cdot N)$. (d) Cached Euler update: all $N$ tokens are advanced in a single tensor op using the latest velocity $\hat{v}_n$ (freshly computed for $n \in \mathcal{A}_i$, cached otherwise).}
\label{fig:system}
\end{figure}

\section{Method}

In this section, we present our proposed method, Heterogeneous Step Allocation (HSA). We first review the preliminaries of video diffusion transformers and flow matching in Section~\ref{sec:preliminaries}. We then introduce the core token partitioning strategy of HSA in Section~\ref{sec:hsa}. To realize this strategy, we propose KV-cache synchronization and a cached Euler update mechanism in Sections~\ref{sec:kv-cache-sync} and~\ref{sec:cached-euler}. Finally, we discuss key implementation details in Section~\ref{sec:implementation}. A visual overview of the system is provided in Figure~\ref{fig:system}.

\subsection{Preliminaries}
\label{sec:preliminaries}

\paragraph{Video DiT inference.}
We consider a video Diffusion Transformer $f_\theta$ that operates in the latent space of a variational autoencoder (VAE). Given a video of $T_v$ frames at resolution $H \times W$, the VAE encodes it into a compact latent tensor of shape $F \times H_l \times W_l$, where $F = (T_v - 1)/4 + 1$, $H_l = H/8$, $W_l = W/8$ in Wan~\cite{wan2025} for example. The DiT then patchifies this latent with a 3D convolution of patch size $[1, 2, 2]$, yielding a flat token sequence $\x \in \mathbb{R}^{N \times d}$ of length $N = F \cdot (H_l/2) \cdot (W_l/2)$.

\paragraph{Flow matching.}
Following~\cite{lipman2023flow,liu2023flow}, the DiT is trained under the flow-matching objective. The forward process is defined as
\begin{equation}
    \x_\sigma = \x_0 + \sigma \boldsymbol{\epsilon}, \quad \boldsymbol{\epsilon} \sim \mathcal{N}(\mathbf{0}, \mathbf{I}),
\end{equation}
where $\sigma \in [0, \sigma_{\max}]$ is the noise level and $\x_0$ denotes the clean latent. The model is trained to predict the velocity field $\vtheta(\x_\sigma, \sigma)$. At inference, a pre-defined schedule $\sigma_0 > \sigma_1 > \cdots > \sigma_T = 0$ induces a discrete ODE solved by the Euler integrator:
\begin{equation}
    \label{eq:euler}
    \x_{\sigma_{i+1}} = \x_{\sigma_i} + \vtheta(\x_{\sigma_i}, \sigma_i)\cdot(\sigma_{i+1} - \sigma_i).
\end{equation}
Standard practice applies Eq.~\eqref{eq:euler} uniformly to all $N$ tokens at every step $i \in \{0, \ldots, T-1\}$, which we refer to as FM in our experiments.

\paragraph{Self-attention with key-value caching.}
Each DiT block contains a self-attention sublayer where queries, keys, and values $(\mathbf{Q}, \mathbf{K}, \mathbf{V}) \in \mathbb{R}^{N \times d_h}$ are computed from the token sequence. The cost of self-attention is $O(N^2 d_h)$ per block. For long video sequences this dominates the per-step compute.

\subsection{Heterogeneous Step Allocation}
\label{sec:hsa}

\paragraph{Token partitioning.}
Let $\Scal = \{1, \ldots, N\}$ be the full set of token indices. HSA partitions $\Scal$ into $K$ disjoint groups
\begin{equation}
    \Scal = \Gcal_1 \cup \Gcal_2 \cup \cdots \cup \Gcal_K, \quad \Gcal_i \cap \Gcal_j = \emptyset \;\; \forall i \neq j,
\end{equation}
where group $\Gcal_k$ is assigned a step budget $s_k \in \mathbb{Z}^+$. Without loss of generality we order groups so that $s_1 \geq s_2 \geq \cdots \geq s_K$, designating $\Gcal_1$ (with budget $s_1 = T$) as the \emph{baseline} group. The effective average step count per token is
\begin{equation}
    \bar{s} = \sum_{k=1}^{K} \frac{|\Gcal_k|}{N} \cdot s_k,
\end{equation}
and the resulting speedup factor relative to the uniform baseline is $T / \bar{s}$.

\paragraph{Divisor constraint.}
To enable aligned, parallel execution across groups, we restrict each group's budget to divisors of $T$:
\begin{equation}
    s_k \mid T \quad \forall k.
\end{equation}
This guarantees that the set of active tokens at any global iteration $i$ is determined by a simple modular condition: group $\Gcal_k$ is \emph{active} at iteration $i$ if and only if $i \bmod (T / s_k) = 0$. Consequently, all tokens active at iteration $i$ can be batched into a single forward pass with no irregular control flow.

\subsection{KV-Cache Synchronization}
\label{sec:kv-cache-sync}

At global iteration $i$, only the subset of \emph{active} tokens $\mathcal{A}_i \subseteq \Scal$ undergoes a forward pass through the transformer blocks. The remaining tokens $\Scal \setminus \mathcal{A}_i$ are \emph{skipped}. To preserve global attention context, we maintain a per-layer KV cache.

Concretely, let $\mathbf{K}^{(l)}, \mathbf{V}^{(l)} \in \mathbb{R}^{N \times d_h}$ denote the full key and value matrices in self-attention layer $l$. We maintain cached copies $\hat{\mathbf{K}}^{(l)}, \hat{\mathbf{V}}^{(l)}$ initialized at the first iteration. At each iteration $i$, we
\begin{enumerate}
    \item \textbf{Compute fresh KV for active tokens.} Run only the active tokens $\x_{\sigma_i}[\mathcal{A}_i]$ through the QKV projections, yielding $\mathbf{K}[\mathcal{A}_i]$ and $\mathbf{V}[\mathcal{A}_i]$.
    \item \textbf{Update cache.} Write the fresh values into the cache: $\hat{\mathbf{K}}^{(l)}[\mathcal{A}_i] \leftarrow \mathbf{K}[\mathcal{A}_i]$, $\hat{\mathbf{V}}^{(l)}[\mathcal{A}_i] \leftarrow \mathbf{V}[\mathcal{A}_i]$. Cached entries for skipped tokens remain unchanged.
    \item \textbf{Full-context attention.} Active tokens compute queries $\mathbf{Q}[\mathcal{A}_i]$ against the \emph{full} cache $(\hat{\mathbf{K}}^{(l)}, \hat{\mathbf{V}}^{(l)})$, attending to both active and cached tokens.
\end{enumerate}
This ensures that every active token retains full global receptive field at each step, with the KV representations of skipped tokens lagging by at most $T / s_k$ iterations.

Because only $\mathbf{Q}[\mathcal{A}_i]$ participates in self-attention, all subsequent per-token computations within the same block---cross-attention and the feed-forward network---likewise operate exclusively on $\mathcal{A}_i$. Inactive tokens produce no intermediate activations and incur no compute in any sublayer at iteration $i$; they are bypassed entirely until their next active iteration.

Concretely, the active set at iteration $i$ is assembled as $\mathcal{A}_i = \bigcup \{\Gcal_k : i \bmod (T/s_k) = 0\}$, and the positional frequencies (RoPE embeddings) are subsampled to match. The per-iteration sequence length seen by the transformer is therefore $|\mathcal{A}_i|$ rather than $N$, reducing self-attention complexity from $O(N^2)$ to $O(|\mathcal{A}_i| \cdot N)$ (active-token queries against the full KV).

\subsection{Cached Euler Update}
\label{sec:cached-euler}

When token $n \in \Scal \setminus \mathcal{A}_i$ is skipped at iteration $i$, its latent $\x_{\sigma_i}[n]$ must still be updated to reflect the noise level $\sigma_{i+1}$ so that it remains coherent with active tokens at the next step. We achieve this via a \emph{cached Euler step} that reuses the velocity predicted at the most recent active iteration for token $n$.

Let $\hat{v}_n = \vtheta(\x_{\sigma_{i_n^*}}[n], \sigma_{i_n^*})$ be the cached velocity from the last active iteration $i_n^* \leq i$. We advance the latent as
\begin{equation}
    \label{eq:cached-update}
    \x_{\sigma_{i+1}}[n] = \x_{\sigma_i}[n] + \hat{v}_n \cdot (\sigma_{i+1} - \sigma_i).
\end{equation}
Equation~\eqref{eq:cached-update} is a standard incremental Euler step that holds the velocity constant between active steps. Applying it recursively over all skipped steps telescopes to a single step from $\sigma_{i_n^*}$ to $\sigma_{i+1}$, so no latent state beyond the running $\x_{\sigma_i}$ needs to be cached.

In practice, we maintain a single cache tensor $\hat{v}_n$ per token, storing the last predicted velocity and updating it whenever token $n$ is active. At each global iteration, all tokens (both active and skipped) are updated via Eq.~\eqref{eq:cached-update} with the shared step $(\sigma_{i+1} - \sigma_i)$, where active tokens use their freshly computed velocity and skipped tokens use $\hat{v}_n$. This unified update rule avoids branching and can be expressed as a single tensor operation over all $N$ tokens.

\subsection{Implementation Details}
\label{sec:implementation}

\paragraph{Token reordering.}
For memory efficiency, we reorder the token sequence so that all non-baseline tokens (i.e., $\bigcup_{k=2}^{K} \Gcal_k$) are placed contiguously at the front of the sequence, followed by baseline tokens $\Gcal_1$. Because the baseline group $\Gcal_1$ is active at every iteration, its KV representations are always freshly computed and never need to be cached. Consequently, the KV cache need only cover the non-baseline prefix of length $N - |\Gcal_1|$, strictly reducing both the cache footprint and the size of index arithmetic in the KV update step. The reordering is applied once before the denoising loop and inverted before unpatchification.

\begin{figure}[t]
    \includegraphics[width=0.33\linewidth]{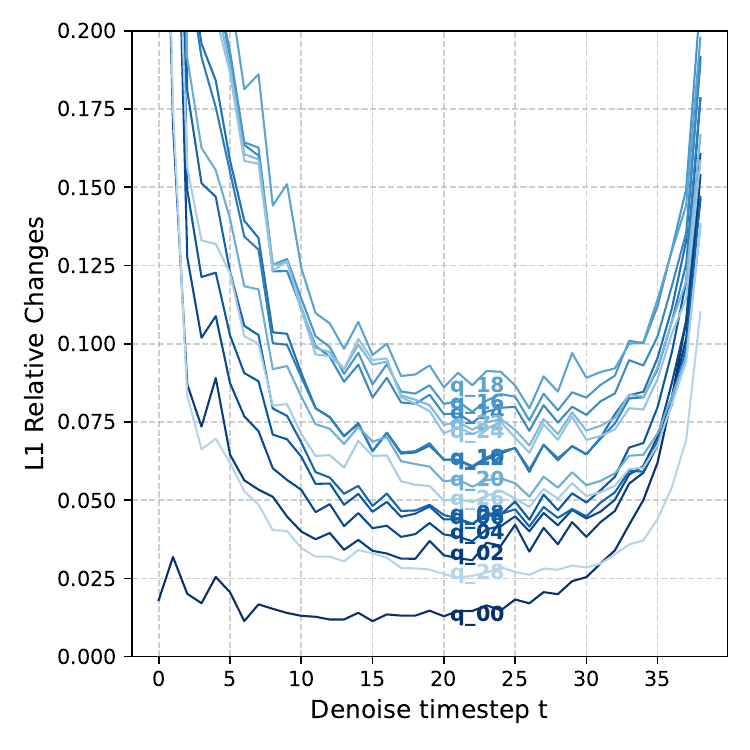}
    \includegraphics[width=0.33\linewidth]{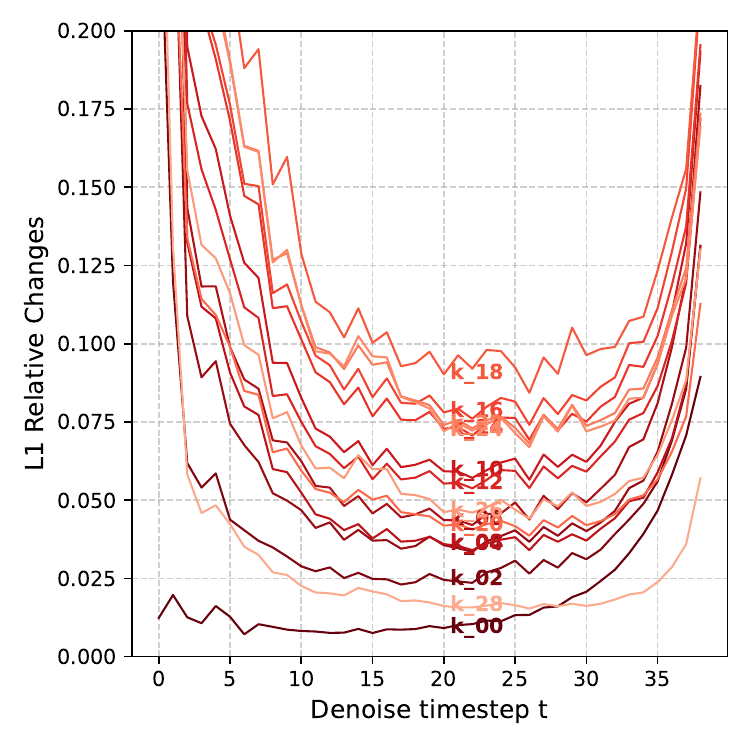}
    \includegraphics[width=0.33\linewidth]{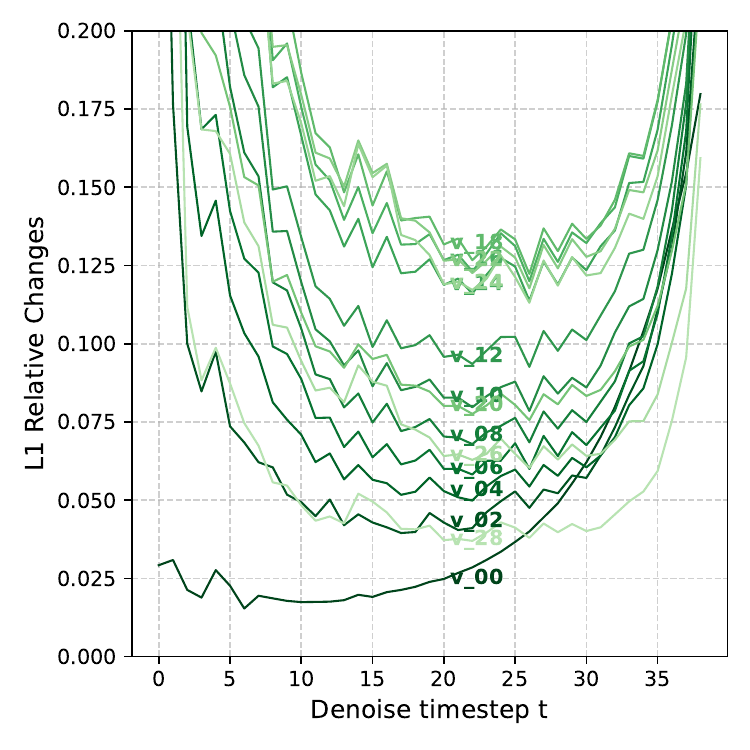}
    \caption{\textbf{L1 relative change~\cite{wimbauer2024cachemeifu} of the QKV vectors for Wan-2.1-1.3B.} Each line tracks the average relative change of the Q/K/V vectors across all tokens in each block at each denoising step, normalized by the average L1 norm of the vectors (e.g. \textit{q\_18} means the query vector in the 18th block). The early and late stages of the trajectory show higher relative change, indicating that they are more sensitive to stale-KV artifacts and motivating our caching window design in Section~\ref{sec:caching-window}.}
    \label{fig:caching-window}
\end{figure}

\paragraph{Phase-aware caching window.}
\label{sec:caching-window}
Prior work~\cite{zhao2024pab,chen2024deltadit,bwcache2025} has shown that the early and late stages of the denoising process carry disproportionate importance: the initial steps establish global structure while the final steps refine fine-grained details, making both phases sensitive to approximation errors, as shown in Figure~\ref{fig:caching-window}. We therefore restrict HSA caching to the \emph{middle} $k\%$ of the denoising trajectory and always execute full, uncached forward passes outside this window. Concretely, let $m = \lfloor (1-k) T / 2 \rfloor$ be the margin in steps. Caching is enabled only for iterations $i \in \{m, m{+}1, \ldots, T{-}m{-}1\}$; the first $m$ and last $m$ steps treat every token as active (i.e., $\mathcal{A}_i = \mathcal{S}$) regardless of group assignment. This window retains the bulk of the computational savings while shielding the quality-critical boundary phases from stale-KV artifacts.

\section{Experiments}

In this section, we evaluate our proposed method. We first introduce the evaluation metrics in Section~\ref{sec:metrics}, detail the evaluation setup in Section~\ref{sec:eval-setup}, then present our main results in Section~\ref{sec:results}.

\subsection{Metrics}
\label{sec:metrics}
We report two complementary families of metrics.

\emph{Distributional quality (primary).}
We use \textbf{VBench}~\cite{huang2024vbench}, the standard benchmark for video generation.
VBench aggregates sixteen sub-dimensions into three headline numbers.
The \textbf{Total Score} is the weighted combination of all dimensions and serves as our primary quality indicator.
It decomposes into a \textbf{Quality Score} (averaging seven per-frame and per-clip dimensions) and a \textbf{Semantic Score} (averaging nine prompt-alignment dimensions).

\emph{Per-sample reference fidelity (secondary).}
We additionally report \textbf{PSNR} and \textbf{LPIPS} of low-budget video samples against the same-seed FM ($T$=40) reference frames, measuring how closely the accelerated trajectory tracks the full-budget one under identical noise. These numbers are informative when the schedule's early-stage denoising---during which the low-frequency global structure is determined---remains close to the reference trajectory; once a schedule perturbs the early stage enough to commit to a different structural basin, the sample can still be perceptually strong and prompt-consistent while scoring poorly against the reference, so a \textbf{drop in PSNR/LPIPS does not necessarily indicate a drop in perceptual quality} (this is why VBench remains the primary reference). We defer the full discussion of when per-sample comparison is meaningful to Appendix~\ref{app:metric-choice}.

\begin{figure}[t]
    \centering
    \begin{subfigure}[t]{0.48\textwidth}
        \centering
        \includegraphics[height=1.2in]{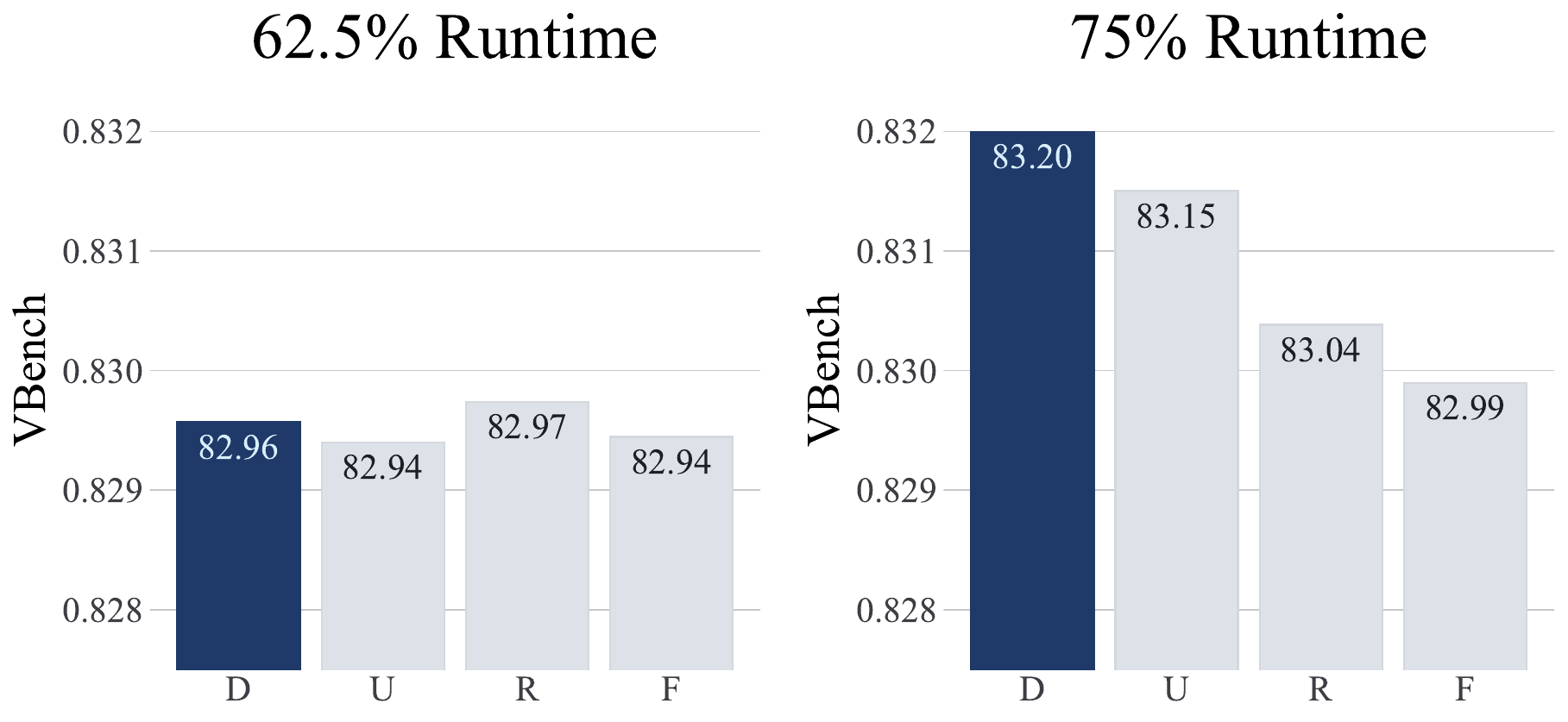}
        \caption{\textbf{Ablation study on token allocation strategy.} VBench T2V score for the four strategies---dynamic (D), uniform (U), random (R), and random with first-frame reservation (F)---at two representative runtime targets ($62.5\%$ and $75\%$).}
        \label{fig:token-alloc-ablation}
    \end{subfigure}%
    \hfill
    \begin{subfigure}[t]{0.48\textwidth}
        \centering
        \includegraphics[height=1.2in]{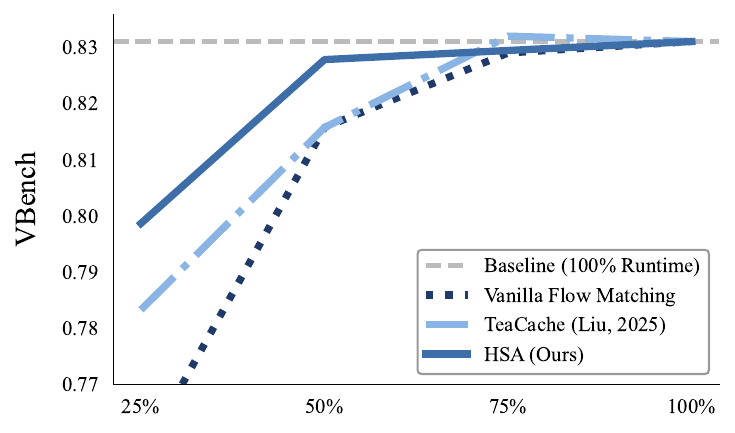}
        \caption{\textbf{Quality-runtime Pareto frontier on Wan-2.1-1.3B.} VBench T2V score as a function of inference cost, measured as a fraction of the $T{=}40$ reference runtime. Our method surpasses vanilla Flow Matching and TeaCache~\cite{liu2024teacache} at 50\% runtime and below, without their expensive offline profiling.}
        \label{fig:pareto-t2v}
    \end{subfigure}
    \caption{\textbf{Results on token allocation strategy and quality-runtime trade-off.}}
\end{figure}

\begin{figure}[t]
    \centering
    \includegraphics[width=\linewidth]{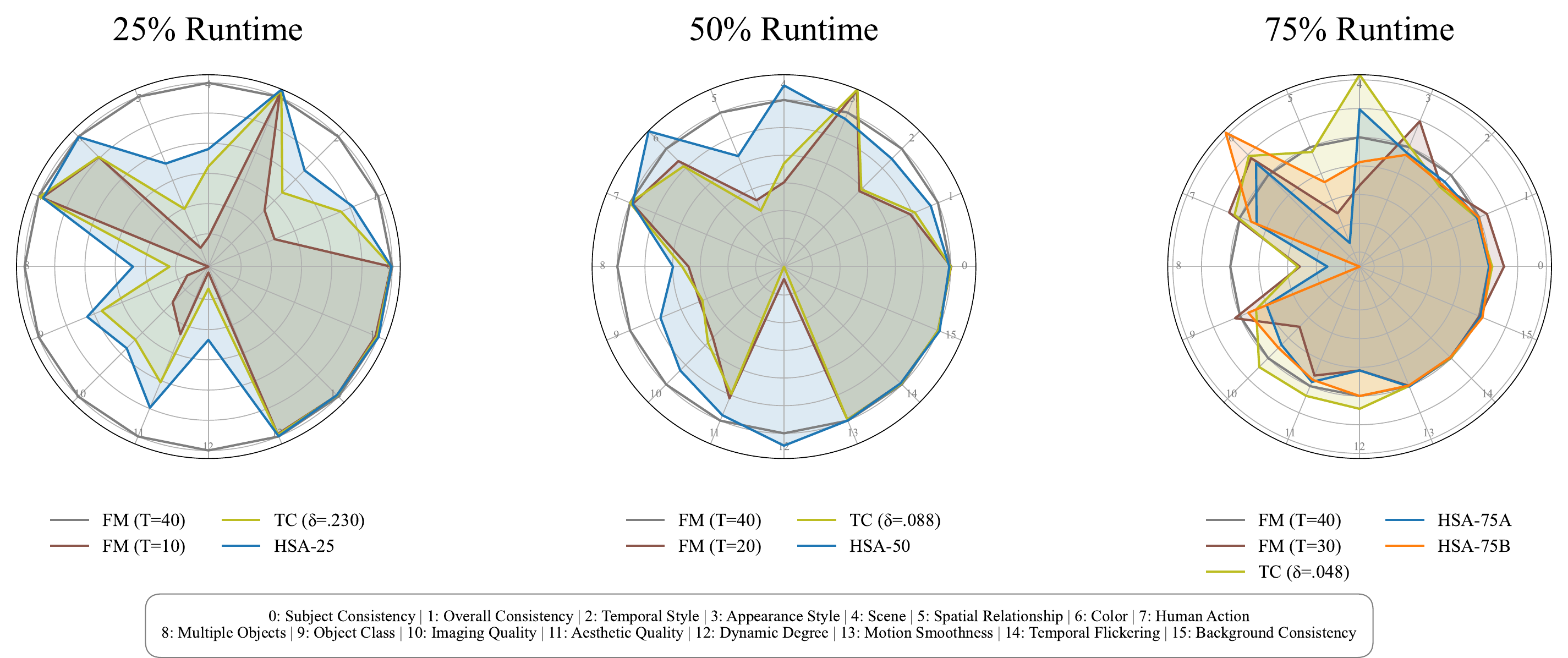}
    \caption{\textbf{Per-dimension VBench profile at different runtime budget on Wan-2.1-1.3B.} Each panel compares HSA against vanilla Flow Matching (FM) at reduced $T$ and TeaCache~\cite{liu2024teacache} (TC) at the same runtime budget; All scores are normalized by the score of full-budget reference FM ($T$=100) for better visualization. HSA better tracks the reference envelope across all sixteen dimensions, while the baselines collapse visibly on several dimensions once the budget is aggressive.}
    \label{fig:radar-t2v}
\end{figure}

\subsection{Evaluation setup}
\label{sec:eval-setup}

Because HSA is a flexible, training-free framework, it can be instantiated with a wide variety of token-grouping strategies and step budget allocations. To comprehensively map this design space, we first sweep across different strategies for assigning tokens to budget groups, and then define a set of representative budget presets spanning various target runtimes. For efficiency, we conduct our primary quantitative evaluations on a smaller model, Wan-2.1-1.3B~\cite{wan2025}, and static qualitative comparison on Wan-2.2-A14B~\cite{wan2025}. To demonstrate the versatility of HSA across model scales, we additionally provide full qualitative comparison on larger models (Wan-2.1-14B/2.2-A14B~\cite{wan2025} and LTX-2~\cite{ltx2026}) on our \href{\supurl}{project page}. We compare against two baselines: \textbf{Flow Matching (FM)}, which applies the same reduced $T$ uniformly to all tokens, and \textbf{TeaCache (TC)}~\cite{liu2024teacache}, a recent state-of-the-art caching method that uses \textbf{time-consuming offline profiling} to estimate model output fluctuations across timesteps. It reuses cached noise prediction when variations are minimal to efficiently reduce redundant computations without sacrificing visual quality.\footnote{We did not compare against other recent methods such as \textbf{HetCache}~\cite{hetcache2026} and \textbf{X-Slim}~\cite{xslim2025} because HetCache targets video editing rather than generation, and X-Slim has not been implemented on the video generators of our interest.}

\paragraph{Token allocation strategies}
We study four strategies for assigning tokens to groups: Dynamic token selection (D), Uniform allocation (U), Random allocation (R), and Random with first-frame reservation (F). Full definitions of these strategies are provided in Appendix~\ref{app:token-alloc}. Before fixing schedules for the main evaluation, we ran a pilot ablation comparing all four strategies at two representative runtime targets, $75\%$ and $62.5\%$ (Fig.~\ref{fig:token-alloc-ablation}). Dynamic selection (D) consistently yields the highest VBench Score.
Consequently, we adopt dynamic selection as the default for the remainder of the paper unless otherwise noted.

\paragraph{Token group presets}
We fix the FM ($T$=40) schedule as the full-budget reference and define four HSA presets that span a range of target runtimes: \textbf{HSA-75A, HSA-75B, HSA-50}, and \textbf{HSA-25}. They are designed to target approximately $75\%$, $75\%$, $50\%$, and $25\%$ runtime to the reference, respectively, with two different presets at the $75\%$ target to demonstrate that HSA can achieve similar runtime-quality trade-offs with different token groupings and budget allocations. Full specifications of the presets can be found in Appendix~\ref{app:presets}.

\begin{table}[t]
\centering
\caption{\textbf{T2V results on Wan-2.1-1.3B.} VBench Total/Quality/Semantic and PSNR/LPIPS to FM ($T$=40) reference. HSA outperforms FM at reduced $T$ and TeaCache~\cite{liu2024teacache} at 50\% runtime and below.}
\label{tab:t2v}
\footnotesize
\begin{tabularx}{\textwidth}{l|C|CCC|CC}
\toprule
\textbf{Scheduler} & \textbf{Runtime $\downarrow$} & \textbf{VBench $\uparrow$} & \textbf{Quality  $\uparrow$} & \textbf{Semantic $\uparrow$} & \textbf{PSNR $\uparrow$} & \textbf{LPIPS $\downarrow$} \\ \midrule
FM ($T$=40) & 100\% & 83.11\% & 83.85\% & 80.13\% & Reference & Reference \\
\midrule
FM ($T$=30) & 75\% & 82.90\% & 83.68\% & 79.79\% & 10.92 $\pm$ 3.19 & 0.60 $\pm$ 0.14 \\
TC ($\delta$=.048)~\cite{liu2024teacache} & 75\% & \textbf{83.20\%} & \textbf{83.98\%} & \textbf{80.08\%} & 26.31 $\pm$ 5.01 & 0.12 $\pm$ 0.07 \\
HSA-75A (Ours) & 75\% & 82.87\% & 83.73\% & 79.43\% & 27.86 $\pm$ 4.13 & 0.10 $\pm$ 0.04 \\
HSA-75B (Ours) & 75\% & 82.95\% & 83.78\% & 79.62\% & 25.82 $\pm$ 4.51 & 0.13 $\pm$ 0.07 \\
\midrule
FM ($T$=20) & 50\% & 81.58\% & 82.58\% & 77.58\% & 14.69 $\pm$ 2.83 & 0.44 $\pm$ 0.10 \\
TC ($\delta$=.088)~\cite{liu2024teacache} & 50\% & 81.58\% & 82.56\% & 77.65\% & 14.73 $\pm$ 2.83 & 0.44 $\pm$ 0.10 \\
HSA-50 (Ours) & 50\% & \textbf{82.79\%} & \textbf{83.66\%} & \textbf{79.30\%} & 21.56 $\pm$ 3.49 & 0.22 $\pm$ 0.08 \\
\midrule
FM ($T$=10) & 25\% & 75.68\% & 77.80\% & 67.20\% & 10.55 $\pm$ 2.41 & 0.65 $\pm$ 0.10 \\
TC ($\delta$=.230)~\cite{liu2024teacache} & 25\% & 78.33\% & 79.86\% & 72.20\% & 10.47 $\pm$ 2.43 & 0.64 $\pm$ 0.10 \\
HSA-25 (Ours) & 25\% & \textbf{79.87\%} & \textbf{81.11\%} & \textbf{74.89\%} & 10.39 $\pm$ 2.37 & 0.64 $\pm$ 0.10 \\
\bottomrule
\end{tabularx}
\end{table}

\subsection{Results}
\label{sec:results}

\paragraph{Text-to-video generation.}
As shown in Table~\ref{tab:t2v}, HSA demonstrates a superior quality-runtime trade-off compared to the baselines. While HSA remains competitive with uniform Flow Matching (FM) and TeaCache (TC) at higher runtime budgets (75\%), its advantages become highly pronounced at aggressive acceleration regimes. At 50\% and 25\% runtimes, HSA significantly outperforms both FM and TC on the VBench benchmark. Figure~\ref{fig:pareto-t2v} illustrates this Pareto frontier, highlighting that HSA maintains higher generation quality without relying on the expensive offline profiling required by TeaCache. Furthermore, the detailed VBench profile in Figure~\ref{fig:radar-t2v} reveals that HSA robustly tracks the full-budget reference envelope across all sixteen evaluation dimensions, whereas the baselines suffer from catastrophic dimension collapse under tight budgets.

\begin{figure}[t]
    \centering
    \begin{tabularx}{\textwidth}{X X X}
        Reference, FM ($T$=40) & FM ($T$=10) & HSA-25 (Ours) \\
    \end{tabularx}
    \includegraphics[width=\linewidth]{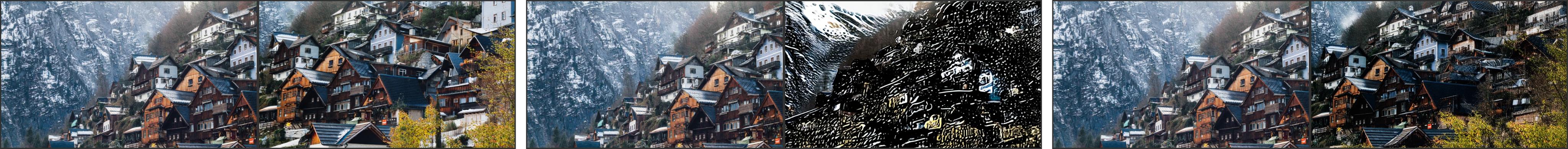}
    \includegraphics[width=\linewidth]{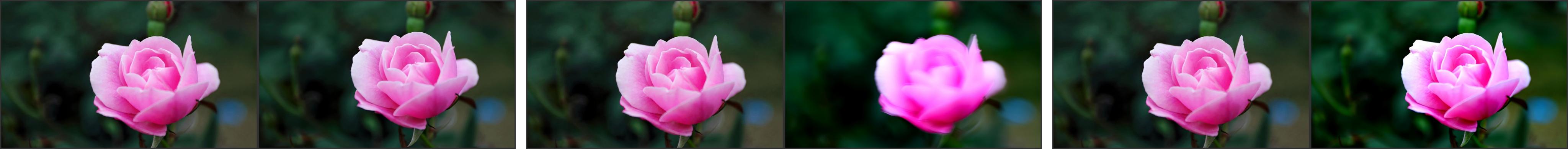}
    \includegraphics[width=\linewidth]{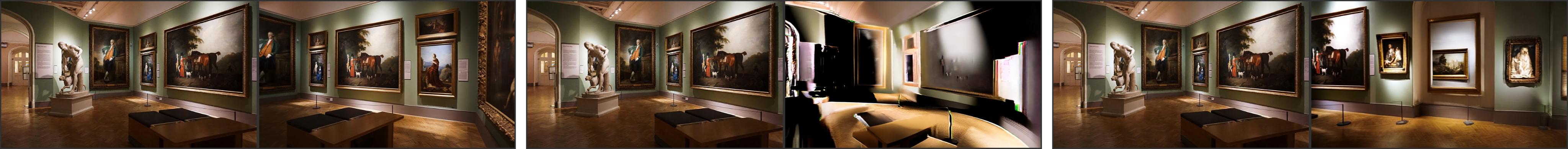}
    \includegraphics[width=\linewidth]{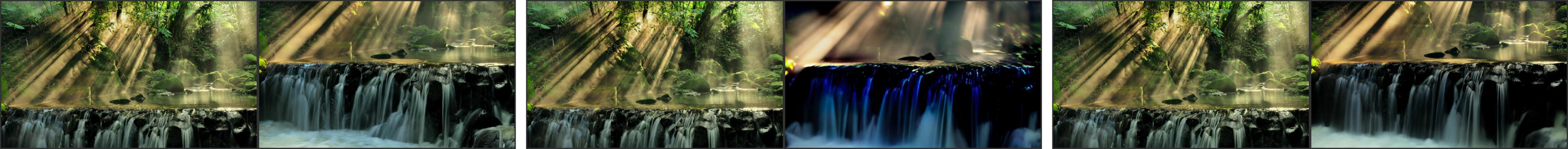}
    \includegraphics[width=\linewidth]{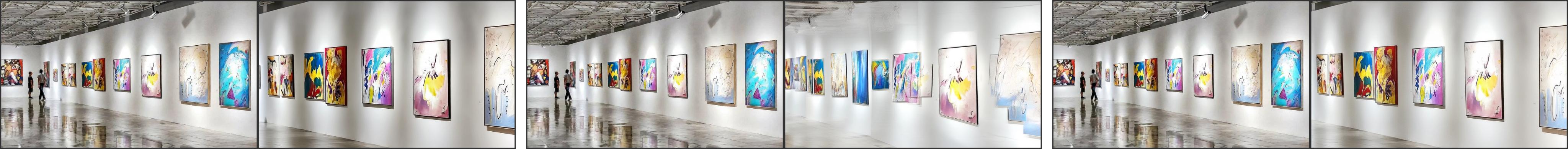}
    \caption{\textbf{Qualitative results for Wan-2.2-A14B image-to-video generation at 25\% runtime.} Each row displays the first and last frames of a video generated using the same image-text prompt across different schedulers. Our proposed HSA-25 successfully preserves strong image-conditioning alignment and high visual fidelity. In contrast, the baseline FM at $T$=10 experiences severe quality degradation and structural misalignment by the final frame. Full videos are available on our \href{\supurl}{website}.}
    \label{fig:wan_a14b_i2v}
\end{figure}

\paragraph{Image-to-video generation.}
We extend our evaluation to image-to-video (I2V) generation, observing similarly strong performance. Figure~\ref{fig:wan_a14b_i2v} showcases generation results on the larger Wan-2.2-A14B model at just 25\% runtime. HSA successfully preserves strong image-conditioning alignment, rich aesthetics, and high visual fidelity throughout the video. In contrast, the uniform FM baseline ($T$=10) experiences severe degradation and structural collapse by the final frame. We also provide the complete visual comparison on our \href{\supurl}{project page}, which includes full videos in Figure~\ref{fig:wan_a14b_i2v} and additional samples on larger models (Wan-2.1-14B/2.2-A14B~\cite{wan2025} and LTX-2~\cite{ltx2026}) across both T2V and I2V.

\section{Related Work}

\paragraph{Step-level feature caching.}
The dominant paradigm for training-free DiT acceleration is to skip entire denoising steps for the whole model and reuse previously computed features in their place.
Wimbauer et al.~\cite{wimbauer2024cachemeifu} introduced block caching with a static L1-based schedule.
TeaCache~\cite{liu2024teacache} makes the decision dynamic by monitoring the L1 change of timestep-embedding-modulated inputs and fitting a polynomial to predict output variation.
MagCache~\cite{magcache2025} discovers that the magnitude ratio of successive residuals follows a prompt-invariant law, enabling single-sample calibration.
EasyCache~\cite{easycache2025} removes offline profiling entirely by tracking a runtime transformation-rate stability criterion.
DiCache~\cite{dicache2025} replaces static priors with an online probe that executes only the first few transformer layers to estimate a per-sample caching indicator.
SenCache~\cite{sencache2026} provides a theoretical grounding: it frames the caching decision as minimizing a first-order sensitivity score composed of Jacobian norms with respect to both the latent and the timestep, unifying TeaCache and MagCache as single-term approximations.
SeaCache~\cite{seacache2026} shifts the decision to the spectral domain, separating structural signal from stochastic noise via a spectral-evolution-aware filter.
OmniCache~\cite{chu2025omnicache} takes a trajectory-global view, concentrating cache reuse at points of minimal curvature and applying adaptive noise correction.
MixCache~\cite{mixcache2025} further generalizes by choosing, at each step, among step-, CFG-, and block-level reuse according to a greedy P-value criterion.

A key property shared by all of these methods is that the caching decision is \emph{globally applied}: at any given iteration, either all tokens are computed or all tokens reuse the cached output.
The granularity of heterogeneity is \emph{temporal} (some steps are computed, others are not), not \emph{spatial} (some tokens are computed, others are not).
HSA introduces a fundamentally different dimension: different tokens are assigned different total step counts, so at each iteration a token-specific subset is active while the rest skip---without discarding the global attention context.

\paragraph{Attention- and block-level caching.}
A parallel body of work targets intra-step redundancy at finer granularity.
TGATE~\cite{zhang2024tgate} caches cross-attention maps after they converge semantically, avoiding re-computation in the fidelity-improving phase.
Pyramid Attention Broadcast (PAB)~\cite{zhao2024pab} exploits the observation that spatial, temporal, and cross-attention exhibit different redundancy periods, broadcasting each at its natural frequency.
$\Delta$-DiT~\cite{chen2024deltadit} caches feature \emph{differences} rather than raw outputs, and adapts front/rear block selection to the denoising stage.
ProfilingDiT~\cite{ma2025profilingdit} uses offline SAM2-guided profiling to identify which blocks attend predominantly to static background vs.\ dynamic foreground, then applies selective reuse only to background-dominant blocks.
BWCache~\cite{bwcache2025} discovers a U-shaped block-feature variation pattern across timesteps and reuses entire block outputs whenever an aggregated L1 indicator falls below a threshold.
TaoCache~\cite{taocache2025} focuses on the late denoising stage, where first-order caching methods fail to preserve fine structure; it models second-order noise deltas to maintain geometric consistency under aggressive skipping.
TaylorSeer~\cite{taylorseer2025} replaces reuse with forecasting: it uses Taylor series expansion on the feature trajectory to predict future block outputs, enabling 5$\times$ speedup without the exponential quality decay that limits direct reuse at large intervals.
Like step-level methods, all of these techniques apply their caching decisions uniformly across the token sequence---the question they ask is ``which block's output should be reused at this step?'' not ``which token should be active at this step?''

\paragraph{Multi-axis and unified caching.}
More recent work combines multiple caching axes within a single framework.
X-Slim~\cite{xslim2025} jointly exploits temporal (step), structural (block), and spatial (token) dimensions via a ``push-then-polish'' dual-threshold controller that switches from aggressive step skipping to lightweight block/token refreshes as accumulated error builds up.
HetCache~\cite{hetcache2026} targets masked video-to-video editing: it divides tokens into generative (inside the edit mask), margin, and context groups, applying a triple-regime scheduler (full, partial, reuse) at the step level and selecting representative context tokens via K-Means for partial-compute steps.
CHAI~\cite{chai2026} goes further by breaking the single-inference boundary, reusing entity-level latents from previous generation runs via a Cross-Inference Cache Attention mechanism.

While X-Slim's spatial component and HetCache's token-level grouping are superficially related to HSA, they differ in a fundamental respect: neither method assigns an \emph{explicit per-token step budget}; each token nominally participates in every iteration and is ``refreshed'' or ``reused'' reactively based on local error indicators.
In X-Slim, the spatial refresh policy makes per-step, per-token decisions driven by observed feature change, so token-level skipping is a reactive consequence of the global error controller rather than a pre-allocated schedule.
In HetCache, the three token groups instead determine \emph{which} tokens are computed during partial-compute steps, but the step-level regime (full, partial, reuse) is decided globally for all tokens at each iteration, and the method is specialized to editing tasks that supply a spatial mask.
HSA, by contrast, assigns different \emph{step budgets} $s_k < T$ to different token groups without requiring any spatial prior.
Tokens assigned fewer steps are systematically bypassed at their inactive iterations, while the KV-cache synchronization mechanism ensures that all active tokens at any iteration still attend over the full $N$-token context.
This combination---heterogeneous per-token step budgets plus full-context attention via KV-cache synchronization---is, to our knowledge, not addressed by any prior work.

\section{Conclusion}

In this paper, we introduced Heterogeneous Step Allocation (HSA), a novel, training-free inference algorithm designed to alleviate the computational bottleneck of Diffusion Transformers (DiTs). Unlike prior global step-caching methods that apply identical denoising schedules to all tokens uniformly, HSA recognizes and exploits the inherent spatial and temporal asymmetry of visual data. By dynamically assigning varying step budgets to different spatiotemporal tokens based on their velocity dynamics, HSA ensures that computational resources are concentrated on the tokens that require more frequent updates, while bypassing those that evolve more slowly.

We tackled the challenges of sequence-length mismatch and latent synchronization with two lightweight mechanisms: KV-cache synchronization, which maintains the full global receptive field for active tokens without computing cross-attention for inactive ones; and a cached Euler update, which reliably advances the latent states of skipped tokens without incurring additional model evaluations. Together, these mechanisms preserve the structural integrity of the generation process while significantly reducing the number of effective token-steps.

Experimental results demonstrate that HSA consistently improves the efficiency-quality Pareto frontier. Even without expensive offline profiling, HSA significantly outperforms existing global step-caching methods and uniform Flow Matching baselines, particularly at aggressive acceleration regimes where others suffer from catastrophic dimension collapse. Furthermore, we believe a promising direction for future research involves exploring more advanced token grouping and allocation strategies to enable the generation of videos that better align with human perception at a significantly reduced budget.

\bibliographystyle{unsrt}
\bibliography{refs}


\appendix
\newpage

\section{On the Choice of Metric and Early-Stage Alignment}
\label{app:metric-choice}

This section elaborates on why we foreground VBench while reporting PSNR and LPIPS only as secondary diagnostics. The central observation is that the interpretability of per-sample reference metrics is governed not by the overall compression ratio but by whether the schedule's \emph{early-stage} denoising---during which the low-frequency global structure is determined---remains close to the reference trajectory.

\paragraph{Per-sample reference metrics measure trajectory alignment, not quality.}
PSNR, SSIM, and LPIPS all compare an accelerated generation against a designated reference---in our case the same-seed uniform-$T$ trajectory. They quantify \emph{how closely the accelerated sample tracks that specific reference draw}. This is a useful quantity when the early-stage denoising of the accelerated schedule is well aligned with the reference: the global structure is established along the same trajectory, so residual drift through the mid and late stages is small and predominantly sub-perceptual (minor texture jitter, sub-pixel shifts, faint luminance offsets). In that regime the metrics are reasonably predictive of perceptual quality and are worth reporting.

\paragraph{The reference trajectory is not privileged.}
The uniform-$T$ baseline for a given seed is one draw from the model's distribution; it has no intrinsic claim to being ``correct.'' A user supplies a prompt, not a seed---they care whether the output is high quality and prompt-aligned, not whether it matches the particular sample that a full-budget run would have drawn from the same noise. Same-seed fidelity is a proxy for ``did we preserve the baseline's computation,'' not for ``is the output good.''

\paragraph{Early-stage alignment governs reference-basin membership.}
The early denoising steps determine the low-frequency content of the sample---global composition, subject layout, coarse color. If a schedule perturbs enough of that early-stage computation, the trajectory commits to a different low-frequency structure and the sample is pulled into a different basin: structurally different, but still a valid generation of the same prompt. Once that happens, per-sample metrics do not so much ``collapse'' as become \emph{incoherent}---two perceptually strong, prompt-consistent samples can exhibit low PSNR and high LPIPS simply because they settled on different plausible compositions. The metrics no longer measure the quantity they are meant to measure. Critically, this is not a property of how much total compression is applied: a schedule that aggressively compresses the \emph{late} stages while leaving the early stages intact can stay in the reference basin, whereas a schedule with mild overall compression that perturbs the early stages can leave it.

\paragraph{Metric sensitivity order.}
Under increasing departure from the reference basin the per-sample metrics degrade in a predictable order:
\begin{itemize}
\item PSNR drops first: any pixel-level shift (even sub-perceptual) contributes.
\item SSIM drops next: sensitive to local structural rearrangement.
\item LPIPS drops last: tolerates low-level texture drift but still penalizes semantic mismatches.
\end{itemize}
None of the three remains interpretable once the samples are in different basins.

\paragraph{Why VBench.}
VBench is a distributional benchmark: it evaluates a model by aggregating per-dimension scores over many prompts, not by comparing individual samples to a reference. The sixteen sub-dimensions collectively capture frame-level quality (temporal consistency, motion smoothness, aesthetic and imaging quality) and semantic fidelity (prompt alignment, object recognition, spatial relations). Because VBench does not assume a privileged reference trajectory, it remains meaningful for every schedule we study, including those whose early-stage trajectory departs from the uniform-$T$ reference.

\paragraph{Reporting convention.}
We report PSNR and LPIPS in all tables for completeness and for direct comparison with the caching literature. We draw quality conclusions from VBench and use PSNR/LPIPS only to diagnose whether a given schedule remains in the reference basin---i.e., whether its early-stage trajectory is close enough to the uniform-$T$ reference for per-sample comparison to be coherent. A drop in PSNR/LPIPS \emph{without} a corresponding drop in VBench should be read as evidence that the sample has drifted into a different but still high-quality basin, not as a quality regression.

\begin{figure}[t]
    \includegraphics[width=\linewidth]{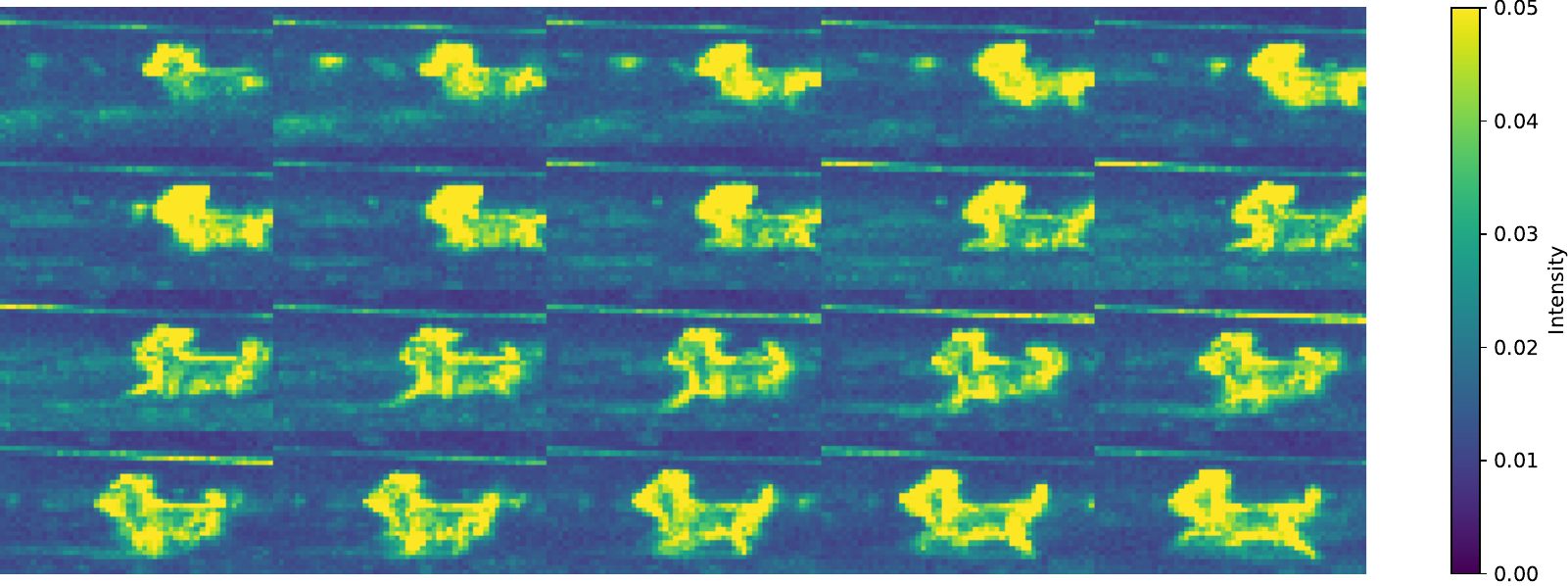}
    \includegraphics[width=0.871\linewidth]{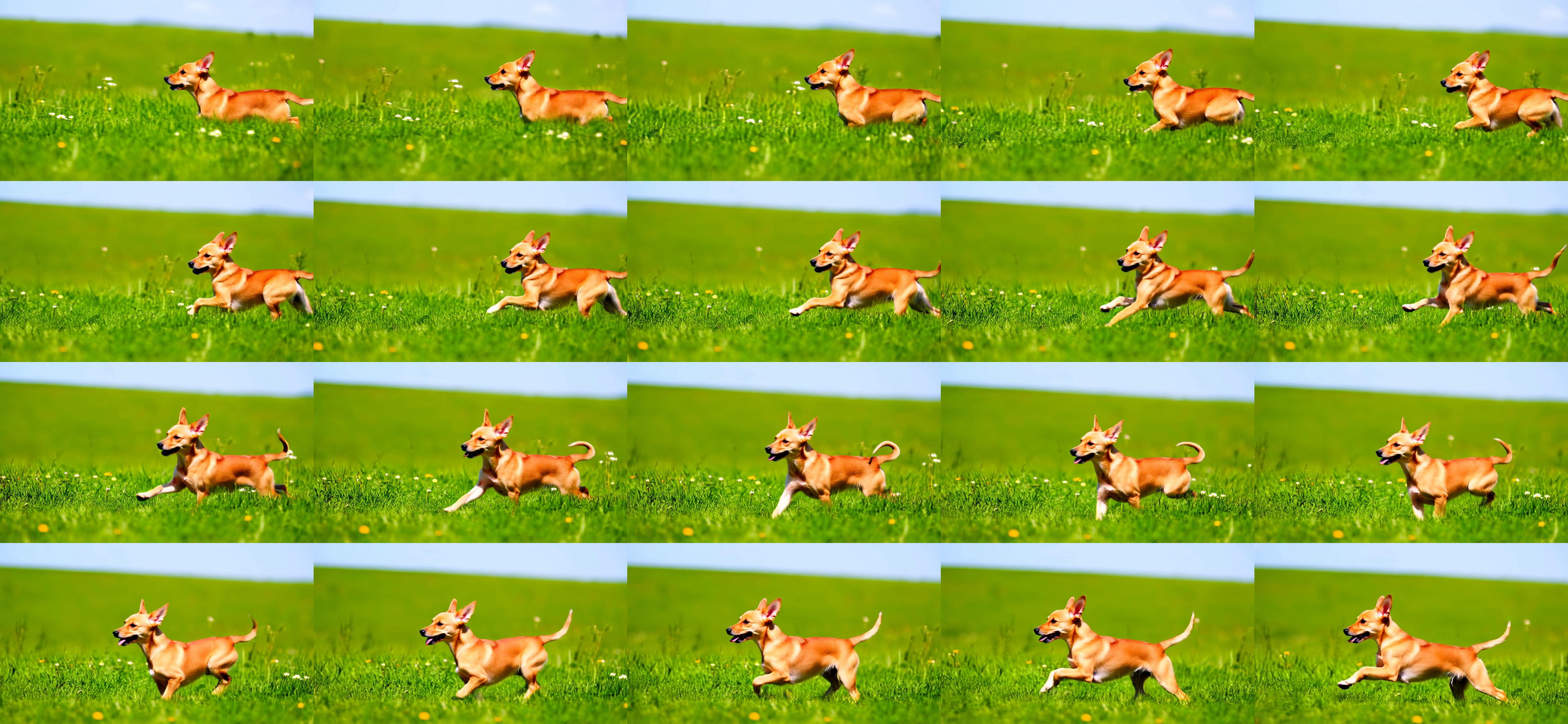}
    \caption{\textbf{Velocity L1 relative change~\cite{wimbauer2024cachemeifu}.} Top: Average per-token L1 relative change of the velocity prediction over the first 5 to 10 steps out of 40 denoising steps. The heatmap indicates that larger relative changes (yellow) are highly localized to the salient subject, while the background exhibits minimal change (blue). Bottom: Corresponding generated image frames. This spatial variance in velocity dynamics is utilized by the dynamic token selection strategy to allocate higher compute budgets to complex regions while aggressively caching the background.}
    \label{fig:vel-l1-rel}
\end{figure}

\section{Token allocation strategies}
\label{app:token-alloc}
\textbf{Dynamic token selection (D)} assigns tokens to groups based on their per-token velocity dynamics rather than position. Inspired by block caching~\cite{wimbauer2024cachemeifu}, at each step, we record the per-token L1 relative change of the velocity prediction,
\begin{equation}
\operatorname{L1}_{\text{rel}}(n, i) = \frac{\|\vtheta(\x_{\sigma_i}[n], \sigma_i) - \vtheta(\x_{\sigma_{i-1}}[n], \sigma_{i-1})\|_1}{\|\vtheta(\x_{\sigma_i}[n], \sigma_i)\|_1},
\end{equation}
where $\vtheta(\x_{\sigma_i}[n], \sigma_i)$ is the velocity prediction for token $n$ at iteration $i$. We then average $\operatorname{L1}_{\text{rel}}(i, t)$ over the initial full-budget steps and rank tokens from smallest to largest change. Tokens with smaller relative changes evolve more slowly and are well approximated by cached velocities, so we assign them to the lower-budget (more aggressively cached) groups, while tokens with larger changes are routed to higher-budget groups up to the baseline $\Gcal_1$. Group sizes $\{|\Gcal_k|\}$ are held fixed to meet target runtimes; only the membership is determined adaptively per sample. A concrete example of the resulting token allocation is visualized in Figure~\ref{fig:vel-l1-rel}.

\textbf{Uniform allocation (U)} places tokens at maximally spread positions within the sequence, approximating a uniform spatial/temporal coverage of each group. Given the row-major token ordering induced by patchification, this prevents any group from concentrating in contiguous spatial or temporal patches, ensuring each group spans the full video volume.

\textbf{Random allocation (R)} draws token indices uniformly at random and distributes them to groups according to the target proportions $\{|\Gcal_k|/N\}$. This is the simplest strategy and requires no structural knowledge of the token sequence.

\textbf{Random with first-frame reservation (F)} is a variant of random allocation that explicitly reserves all tokens corresponding to the first video frame for the baseline group $\Gcal_1$ (i.e., the full $T$-step budget), before distributing remaining tokens randomly. The first latent frame serves as the spatial anchor in image-to-video generation and as a strong conditioning signal even in text-to-video; ensuring it follows the complete denoising trajectory can improve the temporal coherence of the entire sequence.

\section{Token group presets}
\label{app:presets}
We fix the FM ($T$=40) schedule as the full-budget reference and define four HSA presets that span a range of target runtimes. Each preset specifies (i) the group decomposition as the fraction of tokens assigned to each step budget $s_k$, (ii) the phase-aware caching window (Section~\ref{sec:caching-window}), expressed as the central fraction of the denoising trajectory over which reduced-budget groups rely on cached velocities; the remaining early and late steps fall back to the full schedule, and (iii) the token allocation strategy. All presets use the dynamic strategy except \textbf{HSA-25}, who uses the random strategy becuase there are not enough steps before its caching window to properly bootstrap the velocity dynamics. The runtimes of these presets are reported as a percentage of the FM ($T$=40) reference inference time, \textit{i.e.,} $100@40$.

\begin{center}
\small
\begin{tabular}{lclcc}
\toprule
Name & Runtime & Budget allocation (\%tokens@$s_k$) & Caching window & Token allocation strategy \\
\midrule
    \textbf{HSA-75A} & $75\%$   & $25$@$40$ + $25$@$20$ + $50$@$10$     & Center $50\%$ & Dynamic (D) \\
    \textbf{HSA-75B} & $75\%$   & $33$@$40$ + $67$@$10$                 & Center $50\%$ & Dynamic (D) \\
    \textbf{HSA-50}  & $50\%$   & $16$@$40$ + $84$@$10$                 & Center $80\%$ & Dynamic (D) \\
    \textbf{HSA-25}  & $25\%$   & $16$@$20$ + $84$@$5$                  & Center $80\%$ & Random (R) \\
\bottomrule
\end{tabular}
\end{center}

\section{Additional quantitative results}

Table~\ref{tab:t2v-full} provides the complete VBench profile for text-to-video generation on the 1.3B model, covering all sixteen evaluation dimensions. The trends observed in the overall VBench score are reflected across most individual dimensions, with HSA maintaining higher scores than the baselines at reduced runtimes.

Table~\ref{tab:i2v} reports the comprehensive quantitative results for the image-to-video (I2V) task on the 1.3B model. For this task, we additionally report an \textbf{Image-Video Alignment Score (IV-Align.)} covering image-conditioning fidelity (I2V Subject, I2V Background, and Camera Motion)~\cite{huang2025vbench++}.

The I2V task presents an intrinsically easier generation setting compared to text-to-video, as the strong initial image conditioning heavily anchors the generated sequence. Because of this anchoring, most methods stay within a narrow quality band, and absolute metric differences are small. On VBench-I2V, HSA matches or slightly trails the vanilla Flow Matching (FM) baseline at reduced step budgets, reflecting the limited headroom for improvement over the already-anchored reference. The salient takeaway is that HSA successfully reaches the reference quality band while providing a competitive Pareto trade-off between runtime and generation quality. Notably, it adheres much closer to the reference trajectory than prior caching methods, as measured by PSNR and LPIPS.

We place these quantitative results in the appendix because the metrics on the 1.3B model often suffer from low signal-to-noise ratios—the baseline quality itself is limited in this regime, making the numbers less representative of the method's true capability. These quantitative scores do not fully align with the perceptual improvements we observe when HSA is applied to larger models. We strongly encourage readers to consult the qualitative video comparisons generated by the larger 14B models on our \href{\supurl}{supplementary website} for a more accurate assessment of generation quality.

\begin{table}[t]
\centering
\footnotesize 
\caption{\textbf{Full VBench T2V results on Wan-2.1-1.3B for all dimensions.}}
\label{tab:t2v-full}
\begin{tabularx}{\textwidth}{l|CCCC}
\toprule
\multicolumn{1}{c|}{\textbf{Models}} & \textbf{Runtime $\downarrow$} & \textbf{Total Score $\uparrow$} & \textbf{Quality Score$^\dagger$} & \textbf{Semantic Score$^\star$}\\ \midrule
FM ($T$=40) & 100\% & 83.11\% & 83.85\% & 80.13\% \\
\midrule
FM ($T$=30) & 75\% & 82.90\% & 83.68\% & 79.79\% \\
TC ($\delta$=.048)~\cite{liu2024teacache} & 75\% & 83.20\% & 83.98\% & 80.08\% \\
HSA-75A (Ours) & 75\% & 82.87\% & 83.73\% & 79.43\% \\
HSA-75B (Ours) & 75\% & 82.95\% & 83.78\% & 79.62\% \\
\midrule
FM ($T$=20) & 50\% & 81.58\% & 82.58\% & 77.58\% \\
TC ($\delta$=.088)~\cite{liu2024teacache} & 50\% & 81.58\% & 82.56\% & 77.65\% \\
HSA-50 (Ours) & 50\% & 82.79\% & 83.66\% & 79.30\% \\
\midrule
FM ($T$=10) & 25\% & 75.68\% & 77.80\% & 67.20\% \\
TC ($\delta$=.230)~\cite{liu2024teacache} & 25\% & 78.33\% & 79.86\% & 72.20\% \\
HSA-25 (Ours) & 25\% & 79.87\% & 81.11\% & 74.89\% \\
\midrule \midrule
\end{tabularx}
\begin{tabularx}{\textwidth}{l|CCCCCCCC}
\multicolumn{1}{c|}{\textbf{Models}} & \textbf{\begin{tabular}[c]{@{}c@{}}Subject$^\dagger$\\ Consist.\end{tabular}} & \textbf{\begin{tabular}[c]{@{}c@{}}Backgr.$^\dagger$\\ Consist.\end{tabular}} & \textbf{\begin{tabular}[c]{@{}c@{}}Temp.$^\dagger$\\ Flicker.\end{tabular}} & \textbf{\begin{tabular}[c]{@{}c@{}}Motion$^\dagger$\\ Smooth.\end{tabular}} & \textbf{\begin{tabular}[c]{@{}c@{}}Dynamic$^\dagger$\\ Degree\end{tabular}} & \textbf{\begin{tabular}[c]{@{}c@{}}Aesthetic$^\dagger$\\ Quality\end{tabular}} & \textbf{\begin{tabular}[c]{@{}c@{}}Imaging$^\dagger$\\ Quality\end{tabular}} & \textbf{\begin{tabular}[c]{@{}c@{}}Object$^\star$\\ Class\end{tabular}} \\ \midrule
FM ($T$=40) & 92.92\% & 96.82\% & 99.18\% & 98.12\% & 62.22\% & 68.37\% & 65.93\% & 91.46\% \\
FM ($T$=30) & 93.41\% & 96.83\% & 99.15\% & 98.11\% & 61.67\% & 68.10\% & 64.91\% & 91.63\% \\
TC ($\delta$=.048)~\cite{liu2024teacache} & 93.03\% & 96.92\% & 99.16\% & 98.12\% & 62.50\% & 68.62\% & 66.22\% & 90.90\% \\
HSA-75A (Ours) & 92.93\% & 96.87\% & 99.14\% & 98.14\% & 61.67\% & 68.26\% & 65.50\% & 90.54\% \\
HSA-75B (Ours) & 92.97\% & 96.94\% & 99.14\% & 98.10\% & 62.22\% & 68.21\% & 65.60\% & 91.17\% \\
FM ($T$=20) & 92.95\% & 96.85\% & 99.08\% & 98.04\% & 55.28\% & 67.18\% & 62.78\% & 86.46\% \\
TC ($\delta$=.088)~\cite{liu2024teacache} & 92.93\% & 96.84\% & 99.11\% & 98.05\% & 54.72\% & 66.94\% & 63.11\% & 86.27\% \\
HSA-50 (Ours) & 92.83\% & 96.94\% & 99.05\% & 98.12\% & 62.78\% & 68.08\% & 64.98\% & 89.27\% \\
FM ($T$=10) & 92.74\% & 96.37\% & 98.87\% & 97.69\% & 43.89\% & 55.86\% & 51.38\% & 67.09\% \\
TC ($\delta$=.230)~\cite{liu2024teacache} & 92.79\% & 96.62\% & 98.90\% & 97.83\% & 45.56\% & 61.74\% & 57.15\% & 81.11\% \\
HSA-25 (Ours) & 92.80\% & 96.87\% & 98.84\% & 98.09\% & 50.83\% & 64.85\% & 58.47\% & 83.48\% \\
\midrule \midrule
\multicolumn{1}{c|}{\textbf{Models}} & \textbf{\begin{tabular}[c]{@{}c@{}}Multiple$^\star$\\ Objects\end{tabular}} & \textbf{\begin{tabular}[c]{@{}c@{}}Human$^\star$\\ Action\end{tabular}} & \textbf{Color$^\star$} & \textbf{\begin{tabular}[c]{@{}c@{}}Spatial$^\star$\\ Relat.\end{tabular}} & \textbf{Scene$^\star$} & \textbf{\begin{tabular}[c]{@{}c@{}}Appear.$^\star$\\ Style\end{tabular}} & \textbf{\begin{tabular}[c]{@{}c@{}}Temp.$^\star$\\ Style\end{tabular}} & \textbf{\begin{tabular}[c]{@{}c@{}}Overall$^\star$\\ Consist.\end{tabular}} \\ \midrule
FM ($T$=40) & 76.05\% & 96.60\% & 85.96\% & 78.64\% & 55.39\% & 22.85\% & 25.83\% & 26.97\% \\
FM ($T$=30) & 74.21\% & 97.00\% & 86.68\% & 76.68\% & 54.46\% & 23.07\% & 25.70\% & 27.05\% \\
TC ($\delta$=.048)~\cite{liu2024teacache} & 74.25\% & 96.80\% & 86.77\% & 78.50\% & 56.60\% & 22.86\% & 25.69\% & 26.95\% \\
HSA-75A (Ours) & 73.48\% & 96.00\% & 86.47\% & 75.80\% & 55.93\% & 22.80\% & 25.75\% & 26.95\% \\
HSA-75B (Ours) & 72.62\% & 96.20\% & 87.76\% & 77.60\% & 54.91\% & 22.79\% & 25.72\% & 26.97\% \\
FM ($T$=20) & 72.13\% & 96.40\% & 84.87\% & 73.25\% & 52.09\% & 23.25\% & 24.71\% & 26.39\% \\
TC ($\delta$=.088)~\cite{liu2024teacache} & 72.48\% & 96.60\% & 84.42\% & 72.61\% & 52.85\% & 23.27\% & 24.76\% & 26.48\% \\
HSA-50 (Ours) & 72.99\% & 96.40\% & 87.50\% & 75.97\% & 55.97\% & 22.74\% & 25.57\% & 26.82\% \\
FM ($T$=10) & 52.91\% & 96.00\% & 81.86\% & 57.37\% & 41.24\% & 22.96\% & 21.37\% & 21.97\% \\
TC ($\delta$=.230)~\cite{liu2024teacache} & 57.80\% & 96.40\% & 81.90\% & 62.84\% & 47.72\% & 23.11\% & 22.45\% & 25.18\% \\
HSA-25 (Ours) & 62.41\% & 95.80\% & 85.89\% & 69.24\% & 49.32\% & 23.16\% & 23.79\% & 25.76\% \\
\bottomrule
\end{tabularx}
$^\dagger$ Quality dimensions. $^\star$ Semantic dimensions.
\end{table}

\begin{table}[t!]
\centering
\caption{\textbf{I2V results on Wan-2.1-1.3B.} VBench-I2V Total/Quality and image-conditioning alignment (IV-Align., averaging I2V Subject, I2V Background, Camera Motion), plus PSNR/LPIPS to the FM ($T$=40) reference.}
\label{tab:i2v}
\footnotesize 
\begin{tabularx}{\textwidth}{l|C|CCC|CC}
\toprule
\textbf{Scheduler} & \textbf{Runtime $\downarrow$} & \textbf{VBench-I2V} & \textbf{Quality $\uparrow$} & \textbf{IV-Align. $\uparrow$} & \textbf{PSNR $\uparrow$} & \textbf{LPIPS $\downarrow$} \\ \midrule
FM ($T$=40) & 100\% & 89.20\% & 82.94\% & 95.45\% & Reference & Reference \\
\midrule
FM ($T$=30) & 75\% & 89.17\% & 82.90\% & 95.44\% & 25.39 $\pm$ 5.87 & 0.15 $\pm$ 0.10 \\
HSA-75A (Ours) & 75\% & 89.15\% & 82.89\% & 95.41\% & 29.72 $\pm$ 4.54 & 0.08 $\pm$ 0.05 \\
HSA-75B (Ours) & 75\% & 89.14\% & 82.87\% & 95.41\% & 29.73 $\pm$ 4.54 & 0.08 $\pm$ 0.05 \\
\midrule
FM ($T$=20) & 50\% & 89.14\% & 82.86\% & 95.41\% & 22.28 $\pm$ 4.90 & 0.20 $\pm$ 0.11 \\
HSA-50 (Ours) & 50\% & 89.07\% & 82.73\% & 95.41\% & 26.17 $\pm$ 4.39 & 0.12 $\pm$ 0.07 \\
\midrule
FM ($T$=10) & 25\% & 88.92\% & 82.55\% & 95.28\% & 16.82 $\pm$ 3.23 & 0.38 $\pm$ 0.11 \\
HSA-25 (Ours) & 25\% & 88.87\% & 82.39\% & 95.35\% & 16.69 $\pm$ 3.31 & 0.38 $\pm$ 0.11 \\
\bottomrule
\end{tabularx}
\end{table}


\section{Compute resources}
\label{app:compute-resources}
All videos were generated on a server with 8 NVIDIA A5000 GPUs. Each VBench entry takes approximately 1.5 days wall-clock time to compute. When running large models, e.g. Wan-2.1-14B, on 24GB GPUs, we offload the KV cache to CPU memory, thus it does not essentially improve the runtime. However, this limitation does not exist for smaller models or with larger GPU memory.

\section{Broader Impacts, Safeguards, and Licenses}
\label{app:broader-impacts}

The HSA framework offers significant positive societal benefits by democratizing video content creation, providing artists and creators with highly efficient and accessible tools. Conversely, as with many generative AI models, there is an inherent risk of malicious application, particularly in the generation of misleading media such as deepfakes. However, the enhanced inference efficiency of HSA actively reduces computational overhead. This not only mitigates the environmental footprint associated with large-scale video generation but also broadens global access to these advanced capabilities.

To address the potential misuse of video generation technologies, we emphasize the importance of robust safeguards. Our terms of use strictly prohibit the generation of deceptive content, explicitly forbidding the creation of deepfakes for disinformation campaigns. Furthermore, we actively encourage the broader research community to advance the development of reliable detection mechanisms and safety protocols, fostering the responsible deployment of generative AI.

Our implementation leverages foundational components from established models, specifically Wan-2.1/2.2~\cite{wan2025} (distributed under the Apache 2.0 license) and LTX-2~\cite{ltx2026} (distributed under the LTX-2 Community License). The HSA model itself will be released under the CreativeML license. This licensing structure explicitly permits academic and research applications while strictly forbidding the use of the model to generate deceptive or harmful content.

%
%



\end{document}